\documentclass[letterpaper]{article} 
\usepackage{aaai2026}  
\usepackage{times}  
\usepackage{helvet}  
\usepackage{courier}  
\usepackage[hyphens]{url}  
\usepackage{graphicx} 
\usepackage{natbib}  
\usepackage{caption} 
\usepackage{algorithm}
\usepackage{algorithmic}

\usepackage{amsfonts}
\usepackage{amsmath}
\usepackage{booktabs}
\usepackage{multirow} 
\usepackage{tabularx}
\usepackage{adjustbox}
\usepackage{newfloat}
\usepackage{listings}
\DeclareCaptionStyle{ruled}{labelfont=normalfont,labelsep=colon,strut=off} 
\floatstyle{ruled}
\newfloat{listing}{tb}{lst}{}
\floatname{listing}{Listing}
\title{BridgeShape: Latent Diffusion Schrödinger Bridge for 3D Shape Completion}
\author {
    Dequan Kong\textsuperscript{\rm 1},
    Honghua Chen\textsuperscript{\rm 2}\thanks{Corresponding author},
    Zhe Zhu\textsuperscript{\rm 1},
    Mingqiang Wei\textsuperscript{\rm 1}
}
\affiliations {
    \textsuperscript{\rm 1}Nanjing University of Aeronautics and Astronautics\\
    \textsuperscript{\rm 2}School of Data Science, Lingnan University, Hong Kong SAR\\
    \{dqkong, zhuzhe0619, mqwei\}@nuaa.edu.cn, honghuachen@LN.edu.hk
}

\usepackage{bibentry}

\begin{document}

\maketitle

\begin{abstract}
Existing diffusion-based 3D shape completion methods typically use a conditional paradigm, injecting incomplete shape information into the denoising network via deep feature interactions (e.g., concatenation, cross-attention) to guide sampling toward complete shapes, often represented by voxel-based distance functions. However, these approaches fail to explicitly model the optimal global transport path, leading to suboptimal completions. Moreover, performing diffusion directly in voxel space imposes resolution constraints, limiting the generation of fine-grained geometric details.
To address these challenges, we propose BridgeShape, a novel framework for 3D shape completion via latent diffusion Schrödinger bridge. The key innovations lie in two aspects:
(i) BridgeShape formulates shape completion as an optimal transport problem, explicitly modeling the transition between incomplete and complete shapes to ensure a globally coherent transformation.
(ii) We introduce a Depth-Enhanced Vector Quantized Variational Autoencoder (VQ-VAE) to encode 3D shapes into a compact latent space, leveraging self-projected multi-view depth information enriched with strong DINOv2 features to enhance geometric structural perception.
By operating in a compact yet structurally informative latent space, BridgeShape effectively mitigates resolution constraints and enables more efficient and high-fidelity 3D shape completion.
BridgeShape achieves state-of-the-art performance on 3D shape completion benchmarks, demonstrating superior fidelity at higher resolutions and for unseen object classes.
\end{abstract}

\begin{links}
    \link{Code}{https://github.com/kizzyk/BridgeShape}
\end{links}

\begin{figure*}[!t]
    \centering
    \includegraphics[width=0.95\textwidth]{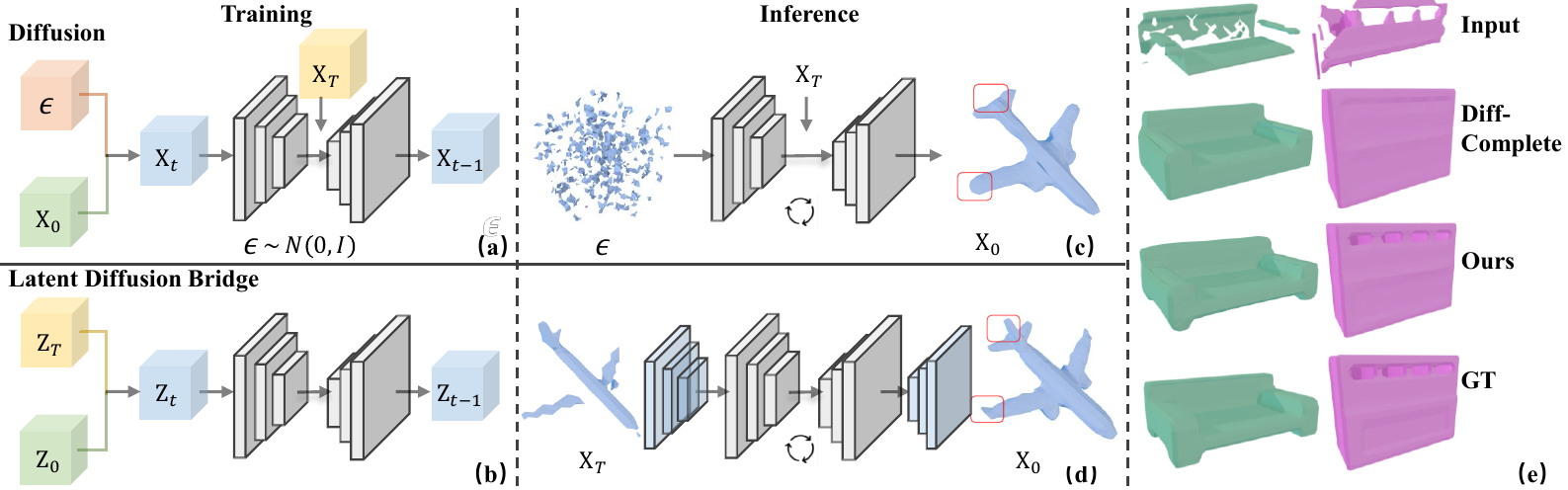}
    \caption{Comparison between existing diffusion-based shape completion paradigms and our proposed latent-diffusion-bridge-based approach. (a) Existing diffusion models incorporate an additional branch to inject deep features into the denoising process, transmitting incomplete shape information without explicitly modeling the transformation between the incomplete shape $\mathrm{X}_T$ and the complete shape $\mathrm{X}_{\mathrm{0}}$. (b) The proposed latent diffusion bridge explicitly models the optimal transport path between the latent distributions of incomplete and complete shapes ($\mathrm{Z}_T$ and $\mathrm{Z}_{\mathrm{0}}$, respectively). (c) Existing diffusion frameworks often produce less coherent completions with missing details, whereas (d) our latent diffusion bridge generates more structurally consistent and detailed 3D shapes. (e) Qualitative comparison of our BridgeShape with DiffComplete~\cite{chu2023diffcomplete}.}
    \label{fig:diffusion}
\end{figure*}

\section{Introduction}
With the rapid advancement of 3D acquisition technologies, 3D sensors have become increasingly accessible and affordable, including various LiDAR scanners deployed across different platforms and RGB-D cameras such as Microsoft Kinect and Intel RealSense. This democratization has significantly expanded their applications across content creation, mixed reality, and machine vision domains \cite{zollhofer2018state, chu2023diffcomplete}. Despite this progress, these sensors are still inherently limited by occlusions, restricted angular fields of view, and surface reflectance issues, often resulting in incomplete and fragmented 3D scans. Such deficiencies hinder downstream tasks that demand complete, high-fidelity 3D representations, thereby necessitating effective shape completion techniques to reconstruct the missing geometric structures in a plausible and accurate manner.

Early approaches to 3D shape completion primarily employ convolutional neural networks \cite{dai2017shape, han2017high} or transformer-based architectures \cite{rao2022patchcomplete} to directly infer complete shapes from partial observations. Building on this, generative models such as variational autoencoders (VAEs) \cite{stutz2020learning, mittal2022autosdf, yan2022shapeformer} and adversarial networks \cite{zhang2021unsupervised, wu2020multimodal, smith2017improved}, were explored to better capture the underlying shape distribution and improve the realism of generated results. More recently, diffusion probabilistic models \cite{sohl2015deep, ho2020denoising} have emerged as a powerful generative framework, achieving state-of-the-art results across various domains  \cite{ramesh2022hierarchical, rombach2022high, nichol2021improved}. Inspired by these successes, recent works~\cite{muller2023diffrf, chou2023diffusionsdf, chu2023diffcomplete, cheng2023sdfusion} have adapted diffusion models for 3D shape completion, adopting a conditional generation paradigm where incomplete shapes are injected into denoising networks via deep feature interactions (e.g., concatenation and cross-attention) to guide sampling toward complete shapes. However, most of these methods operate diffusion directly in voxel space, inherently limiting the resolution of generated shapes and restricting the ability to capture fine-grained geometric details.

Beyond the resolution bottleneck, a more fundamental limitation of diffusion-based methods is that their reverse processes always begin from Gaussian noise, which carries minimal information about the target distribution. As a result, the substantial discrepancy between the Gaussian prior and the target shape distribution necessitates additional conditioning mechanisms and more sampling steps to converge to the target distribution.  However, such implicit conditioning strategies do not explicitly model the global optimal transport path from the incomplete to the complete shape distribution, leading to suboptimal reconstructions and loss of geometric fidelity (see Fig.~\ref{fig:diffusion} (c) and (e)). 

To address these limitations, we introduce a new paradigm for 3D shape completion. Given that the task inherently involves transforming an incomplete shape into a plausible complete one, it is intuitive to start the generative process from the given partial input—rather than from noise—which provides a far more informative and structured prior. This insight motivates the use of diffusion bridge models \cite{li2023bbdm, i2sb}, which condition the diffusion process on both the starting (incomplete) and target (complete) distributions, thereby learning an explicit optimal transport path. While recent works such as I$^2$SB \cite{i2sb} have demonstrated the effectiveness of this idea in image domains, their potential for 3D shape completion remains largely unexplored.

We propose BridgeShape, a novel framework for 3D shape completion via the latent diffusion Schrödinger bridge~(DSB). First, to address the computational challenges inherent in 3D representations, we introduce a Depth-Enhanced VQ-VAE \cite{van2017neural} that compresses 3D shapes into a compact yet structurally informative latent space. This representation integrates self-projected multi-view depth information enriched with strong DINOv2 \cite{oquab2023dinov2} features, enhancing its geometric structural perception. Then, within the latent space, we formulate the DSB to explicitly model the optimal transport between the distributions of partial and complete shapes, enabling a globally consistent and detail-preserving transformation. By leveraging the structurally informative latent encoding and the optimality of the Schrödinger bridge formulation, BridgeShape achieves fine-grained, high-fidelity completions even under challenging conditions. Extensive experiments on large-scale 3D shape completion benchmarks demonstrate state-of-the-art performance, even when generalizing to unseen object categories and higher resolutions.
Our main contributions are summarized as follows:
\begin{itemize}
   \item We propose BridgeShape, casting 3D shape completion as an optimal‑transport problem in latent space.
   \item We introduce a Depth‑Enhanced VQ‑VAE that encodes 3D shapes into a latent space, leveraging self-projected multi-view depth to enhance geometric perception.
   \item On large-scale 3D completion benchmarks, BridgeShape achieves impressive performance even at higher resolutions and on unseen object classes.
\end{itemize}

\label{sec:intro}

\section{Related Work}
 
\subsection{3D Shape Completion} 
3D shape completion recovers missing regions in scans, crucial for scene understanding. Early learning-based methods~\cite{dai2017shape, han2017high} rely on convolutional neural networks, while more recent approaches leverage transformer-based architectures. For instance, 3D-EPN~\cite{dai2017shape} employs a 3D encoder-decoder framework to infer complete shapes, and PatchComplete~\cite{rao2022patchcomplete} leverages multi-resolution patch priors for generalization. A series of point-cloud-based approaches~\cite{pmp, li2025genpc, yu2022adapointr, SVDFormer, CSDN, pointsea} have also addressed this task.  
Generative models—GANs~\cite{zhang2021unsupervised, wu2020multimodal, smith2017improved} and autoencoders~\cite{mittal2022autosdf, yan2022shapeformer}—model completion uncertainty. cGAN~\cite{wu2020multimodal} distills the ambiguity by conditioning the completion on a learned multimodal distribution, while ShapeFormer~\cite{yan2022shapeformer} generates complete sequences.
In contrast, we propose a novel framework that explicitly transports incomplete to complete shapes in a compact, depth-enhanced latent space, ensuring coherent and precise results.  

\subsection{Diffusion models for 3D generation} 
Diffusion models~\cite{song2020score, rombach2022high, dhariwal2021diffusion, sohl2015deep, shim2023diffusion, wang2024mvdd} have proven powerful for 3D shape generation and have recently been adapted to point cloud synthesis~\cite{pvd, luo2021diffusion, vahdat2022lion}.  
~
For conditional shape completion, SDFusion~\cite{cheng2023sdfusion} and Diffusion-SDF~\cite{chou2023diffusionsdf} use diffusion to fill missing regions on synthetically cropped shapes. In contrast, our approach makes no strict assumptions on partial inputs, handling diverse noise and incompleteness.
~
Another approach, DiffComplete~\cite{chu2023diffcomplete}, adopts a multi-level aggregation strategy to improve completion quality. However, performing diffusion directly in voxel space requires substantial memory at higher resolutions. While scaling techniques such as gradient accumulation can mitigate this, they incur extra training complexity.
Moreover, a common limitation of these methods is their inability to explicitly model the optimal transport path between incomplete and complete shapes, often leading to suboptimal completions. BridgeShape overcomes this by directly modeling the optimal transport process within a compact latent space, thereby enabling higher-fidelity completions while alleviating resolution constraints.

\subsection{Diffusion Bridge Models}  
Stochastic bridge models, which capture the evolution of stochastic processes constrained by fixed endpoints, have become an essential tool in probability theory~\cite{aguilar2022sampling, chen2021stochastic, chen2015stochastic}. Integrated with diffusion models, they offer a novel approach for translating between distributions and modeling conditional probabilities without relying on prior information. This data-to-data generation paradigm has attracted significant interest in various generative tasks, including image-to-image translation~\cite{de2021diffusion, chen2021likelihood, li2023bbdm, i2sb}, protein matching~\cite{somnath2023aligned}, point cloud denoising~\cite{vogel2024p2p}, and text-to-speech synthesis~\cite{chen2023schrodinger}. However, the application of diffusion bridge models to 3D shape completion remains unexplored. 

\label{sec:related}

\begin{figure*}[!t] 
\centering 
\includegraphics[width=0.95\textwidth]{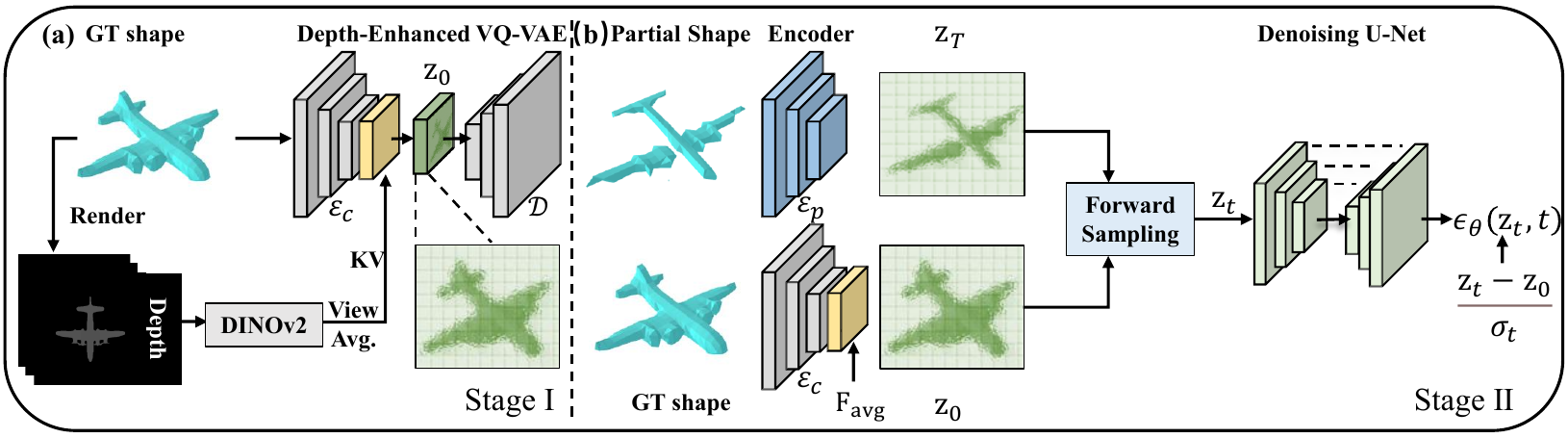} 
\caption{Overview of our training pipeline, which operates within the latent space based on the DSB. \textbf{Stage I}: Pre-training a Depth-Enhanced VQ-VAE on complete shapes to establish the latent space. \textbf{Stage II}: A co-trained encoder maps partial TSDF inputs into this latent space, where the diffusion bridge is applied to learn a structured diffusion trajectory between incomplete and complete shapes. This approach significantly enhances both efficiency and fidelity in shape completion.} 
\label{fig:pipline} 
\end{figure*}

\section{Method}
\subsection{Overview}
The architecture of our approach is shown in Fig.~\ref{fig:pipline}. 
Given a partial scan and its corresponding ground truth complete shape, following previous practice~\cite{chu2023diffcomplete}, we represent the partial scan as a truncated signed distance field (TSDF) and the complete shape as a truncated unsigned distance field (TUDF) in a volumetric grid. 
To accelerate the diffusion-based completion framework, we first compress the complete shape into a low-resolution latent space using a VQ-VAE~\cite{razavi2019generating,van2017neural,cheng2023sdfusion} enhanced by multi-view depth information.
~
Built upon the pretrained latent space, the DSB~\cite{i2sb} is utilized to model the diffusion process from the partial to the complete shape.

\subsection{3D Shape Compression}
\label{sec:3D Shape Compression}
\subsubsection{VQ-VAE Compression.}

The VQ-VAE has an encoder $\mathcal{E}_c$ mapping a 3D shape into a latent space and a decoder $\mathcal{D}$ reconstructing it from the latent code. This architecture enables the application of diffusion models by providing a compact, lower-dimensional representation of the shape.

Let $ \mathbf{X} \in \mathbb{R}^{D \times D \times D}$ represent a complete 3D shape in the form of a TUDF. The encoder $\mathcal{  E }_c$ maps this shape to a latent vector $ \mathbf{z} \in \mathbb{R}^{d \times d \times d} $, where $ d < D $: \begin{equation}
\mathbf{z} = \mathcal{E}_c(\mathbf{X}),
\end{equation} 
the latent vector $ \mathbf{z} $ is then quantized by selecting the closest codebook element $ Z $: 
\begin{equation} 
\mathcal{VQ}(\mathbf{z}) = \arg\min_{\mathbf{z}_i \in Z} \|\mathbf{z} - \mathbf{z}_i\|_2,
\end{equation} where $ \mathbf{z}_i \in Z $ represents an element from the codebook, and $ \|\cdot\|_2 $ denotes the L2 norm. The decoder $\mathcal{ D }$ reconstructs the shape from $\mathbf{z}$ defined as: \begin{equation} \mathbf{X}^{\prime} = \mathcal{ D }(\mathcal{VQ}(\mathbf{z})). \end{equation}

The encoder, decoder, and codebook are jointly optimized with the following total loss function: 
\begin{equation}
\label{eq:vqloss}
\mathcal{L}_{\mathrm{total}}=-\log p(\mathbf{X} \mid \mathbf{z}) +\|\hat{\mathbf{z}}-\operatorname{sg}[\mathbf{z}]\|^{2} + 
  \|\operatorname{sg}[\hat{\mathbf{z}}]-\mathbf{z}\|^{2},
\end{equation}
where $\text{sg}[\cdot]$ is the stop-gradient operation. The three terms in Equation~\ref{eq:vqloss} are: the reconstruction loss, the commitment loss, and the VQ objective.

\subsubsection{Enhancement with Multi-View Depth Features.}

To strengthen the latent space representation, we incorporate multi-view depth features into the VQ-VAE. Specifically, we project the 3D shape into depth maps from \(N\) viewpoints, where each depth map is denoted as \(\mathbf{D}_i\) for the \(i\)-th view, and its corresponding feature \(\mathbf{F}_i\) is extracted using a pre-trained DINOv2 model \cite{oquab2023dinov2}. To obtain a unified representation, we aggregate these features by averaging along the view dimension as a simple yet effective fusion strategy:
\begin{equation}
\mathbf{F}_{\text{avg}} = \frac{1}{N} \sum_{i=1}^{N} \mathbf{F}_i.
\end{equation}
This operation produces an aggregated feature map \(\mathbf{F}_{\text{avg}}\) that fuses complementary information across all viewpoints while minimizing redundant details, resulting in a robust representation with minimal computational overhead.
~
\(\mathbf{F}_{\text{avg}}\) is then fused with the 3D shape's latent feature via a cross-attention mechanism at the end of the encoder~$\mathcal{  E }_c$.  In this fusion process, the query matrix is derived from the 3D shape's latent feature, while the key and value matrix are derived from $\mathbf{F}_{\text{avg}}$. This fusion is performed at a critical stage—the end of the encoder—where the shape information has already been compressed into a high-level representation.  
This approach strikes a balance between model performance and resource utilization, avoiding unnecessary overhead while enriching the latent space representation.

\subsection{Latent DSB}
\label{sec:Latent Diffusion Bridge}

After compressing complete 3D shapes into a high-dimensional latent space, we construct a DSB to model the optimal transport between incomplete and complete shapes. 
As illustrated in Figure~\ref{fig:pipline}(b), we freeze the pre-trained VQ-VAE parameters and employ an additional trainable encoder, \( \mathcal{E}_p \), to map partial inputs into the same latent space, where the latent representations of a given incomplete shape and its corresponding complete shape are denoted by \( \mathbf{z}_T \) and \( \mathbf{z}_0 \), respectively. Then, the diffusion bridge enables efficient processing within this compact yet structurally informative space. 
In the following, we detail how 3D shape completion is formulated as an optimal transport problem via the diffusion bridge and provide an overview of the associated training and inference procedures.

\subsubsection{Optimal Transport via DSB.} DSB is a specialized diffusion process that progressively transforms \(\mathbf{z}_0 \sim p_{\text{A}}\) into \(\mathbf{z}_T \sim p_{\text{B}}\) via a sequence of intermediate states \(\{\mathbf{z}_1, \dots, \mathbf{z}_T\}\) over \(T\) timesteps.

Given a reference path measure \(\pi(\mathbf{z}_{0:T})\) that characterizes the ideal diffusion trajectory, our goal is to learn a process \(p^*(\mathbf{z}_{0:T})\) that satisfies \(p^*(\mathbf{z}_0) = p_{\text{A}}\) and \(p^*(\mathbf{z}_T) = p_{\text{B}}\), while minimizing the Kullback-Leibler divergence between \(\pi(\mathbf{z}_{0:T})\) and \(p^*(\mathbf{z}_{0:T})\). This formulation is equivalent to the Schrödinger Bridge problem~\cite{leonard2013survey, chen2023schrodinger, vogel2024p2p}, whose dynamics are characterized by the following stochastic differential equations (SDEs):\begin{equation}
    \text{d} \mathbf{z}_t = [\mathbf{f}(\mathbf{z}_t, t) + g^2(t) \nabla \log{\Psi_t(\mathbf{z}_t)}] \, \text{d}t + g(t) \, \text{d}\mathbf{w}_t, \\
\end{equation}
\begin{equation}
    \text{d} \mathbf{z}_t = [\mathbf{f}(\mathbf{z}_t, t) - g^2(t) \nabla \log{\hat{\Psi}_t(\mathbf{z}_t)}] \, \text{d}t + g(t) \, \text{d}\mathbf{\bar{w}}_t, 
\end{equation}
where $ \mathbf{f}(\mathbf{z}_t, t) $ is the drift term, $ g(t) $ is the diffusion coefficient and $\mathbf{w}_t$ is a Wiener process. The terms $ \nabla \log \Psi_t(\mathbf{z}_t) $ and $ \nabla \log \hat{\Psi}_t(\mathbf{z}_t) $ represent the extra nonlinear drift terms that solve the coupled partial differential equations (PDEs):
\begin{align}
\begin{split}
\label{eq:pdes}
    \begin{cases}
\frac{\partial \Psi}{\partial t} = - \nabla \Psi^{\text{T}} \mathbf{f} - \frac{1}{2} \text{Tr}(g^2 \nabla^2 \Psi) \\
\frac{\partial \hat{\Psi}}{\partial t} = - \nabla (\hat{\Psi} \mathbf{f}) + \frac{1}{2} \text{Tr}(g^2 \nabla^2 \hat{\Psi})
    \end{cases}
\end{split}
\end{align}
such that
\begin{equation}
    \Psi_0 \hat{\Psi}_0 = p_{\text{A}}, \quad \Psi_T \hat{\Psi}_T = p_{\text{B}}.
\end{equation}
However, directly solving the Schrödinger bridge formulation is computationally prohibitive. To mitigate it, we leverage a framework from recent works~\cite{i2sb, chen2023schrodinger, vogel2024p2p} under the assumption that paired training data is available, i.e.,
\begin{equation}
p(\mathbf{z}_0, \mathbf{z}_T) = p_{A}(\mathbf{z}_0)\, p_{B}(\mathbf{z}_T \mid \mathbf{z}_0).
\end{equation}
In the context of 3D shape completion, the latent distribution over incomplete shapes is modeled by a joint distribution: \(p_{A}(\mathbf{z}_0)\) represents the latent distribution of complete shapes, and \(p_{B}(\mathbf{z}_T \mid \mathbf{z}_0)\) characterizes the latent distribution of missing components conditioned on the complete shapes. 

Given a boundary pair \( \mathbf{z}_0 \) (complete shape) and \( \mathbf{z}_T \) (incomplete shape), and assuming \(\mathbf{f}:= 0\), the posterior distribution at each timestep \( t \) is defined as:
\begin{equation}
\label{eq:posterior}
q(\mathbf{z}_t \mid \mathbf{z}_0, \mathbf{z}_T) = \mathcal{N}\left(\mathbf{z}_t; \mu_t(\mathbf{z}_0, \mathbf{z}_T), \Sigma_t\right),
\end{equation}
where the mean \( \mu_t \) and covariance \( \Sigma_t \) are computed as:
\begin{equation}
\mu_t(\mathbf{z}_0, \mathbf{z}_T) = \frac{\sigma_{b,t}^2}{\sigma_{b,t}^2 + \sigma_t^2}\mathbf{z}_0 + \frac{\sigma_t^2}{\sigma_{b,t}^2 + \sigma_t^2}\mathbf{z}_T,
\end{equation}
\begin{equation}
\Sigma_t = \frac{\sigma_t^2\, \sigma_{b,t}^2}{\sigma_t^2 + \sigma_{b,t}^2}.
\end{equation}
Here, \( \sigma_t^2 \) and \( \sigma_{b,t}^2 \) represent the accumulated variance from \( \mathbf{z}_T \) and \( \mathbf{z}_0 \), respectively. This framework enables efficient computation of the coupled PDEs (see Equation~\ref{eq:pdes}), thereby facilitating the latent diffusion process. Moreover, the sampling mechanism in Equation~\ref{eq:posterior} is both tractable and sufficiently expressive to cover the generative trajectory, resulting in an effective and efficient shape completion pipeline.

Considering that extremely sparse incomplete shapes can introduce significant uncertainty in the missing regions, establishing a robust optimal transport path may become particularly challenging. To address this, we inject stochasticity into the latent distribution of incomplete shapes before constructing the optimal transport process—following the strategy employed in \cite{i2sb}. 

\begin{table*}[!t]
\centering
\setlength{\tabcolsep}{1.2mm}
 \begin{tabular*}{\linewidth}{l  c c c c c c c c c }
\toprule 
$l_1$-err.~$\downarrow$  & Chair & Table & Sofa & Lamp & Plane & Car & Dresser & Watercraft & Avg.\\  
\specialrule{0em}{2pt}{0pt}
\hline
\specialrule{0em}{1.5pt}{0pt}
3D-EPN~\cite{dai2017shape} & 0.418 & 0.377 & 0.392 & 0.388 & 0.421 & 0.259 & 0.381 & 0.356 & 0.374 \\
ConvONet~\cite{peng2020convolutional}  & 0.210 & 0.247 & 0.254 & 0.234 & 0.185 & 0.195 & 0.250 & 0.184  & 0.220\\
SDF-StyleGAN~\cite{zheng2022sdf} & 0.321 & 0.256 & 0.289 & 0.280 & 0.295 & 0.224 & 0.273 & 0.282  & 0.278\\ 
cGCA~\cite{zhang2022probabilistic} & 0.174 & 0.212 & 0.179 & 0.239 & 0.170 & 0.161 & 0.204 & 0.143 & 0.185\\ 
AutoSDF~\cite{mittal2022autosdf}  & 0.201 & 0.258 & 0.226 & 0.275 & 0.184 & 0.187 & 0.248 & 0.157 & 0.217\\ 

ShapeFormer~\cite{yan2022shapeformer}  & 0.104 & 0.175 & 0.133 & 0.176 & 0.136 & 0.127 & 0.157 & 0.119 & 0.141\\ 
PatchComplete~\cite{rao2022patchcomplete}  & 0.134 & 0.095 & 0.084 & 0.087 & 0.061 & 0.053 & 0.134 & 0.058 & 0.088\\
Diffcomplete~\cite{chu2023diffcomplete}   & 0.070 & 0.073 & 0.061 & 0.059 & 0.015 & 0.025 & 0.086 & 0.031 & 0.053\\
\specialrule{0em}{1.5pt}{0pt}
\hline
\specialrule{0em}{2pt}{0pt}
BrigdeShape~\small{(Ours)}  & \textbf{0.055} & \textbf{0.059} & \textbf{0.047} & \textbf{0.038} & \textbf{0.012} & \textbf{0.023} & \textbf{0.055} & \textbf{0.022} & \textbf{0.039} \\
\bottomrule
\end{tabular*}
\caption{Quantitative comparison for shape completion on 3D-EPN~\cite{dai2017shape}.}
\label{tab:epn_bench}
\end{table*}

\begin{figure*}[!t]
\centering
\includegraphics[width=0.94\textwidth]{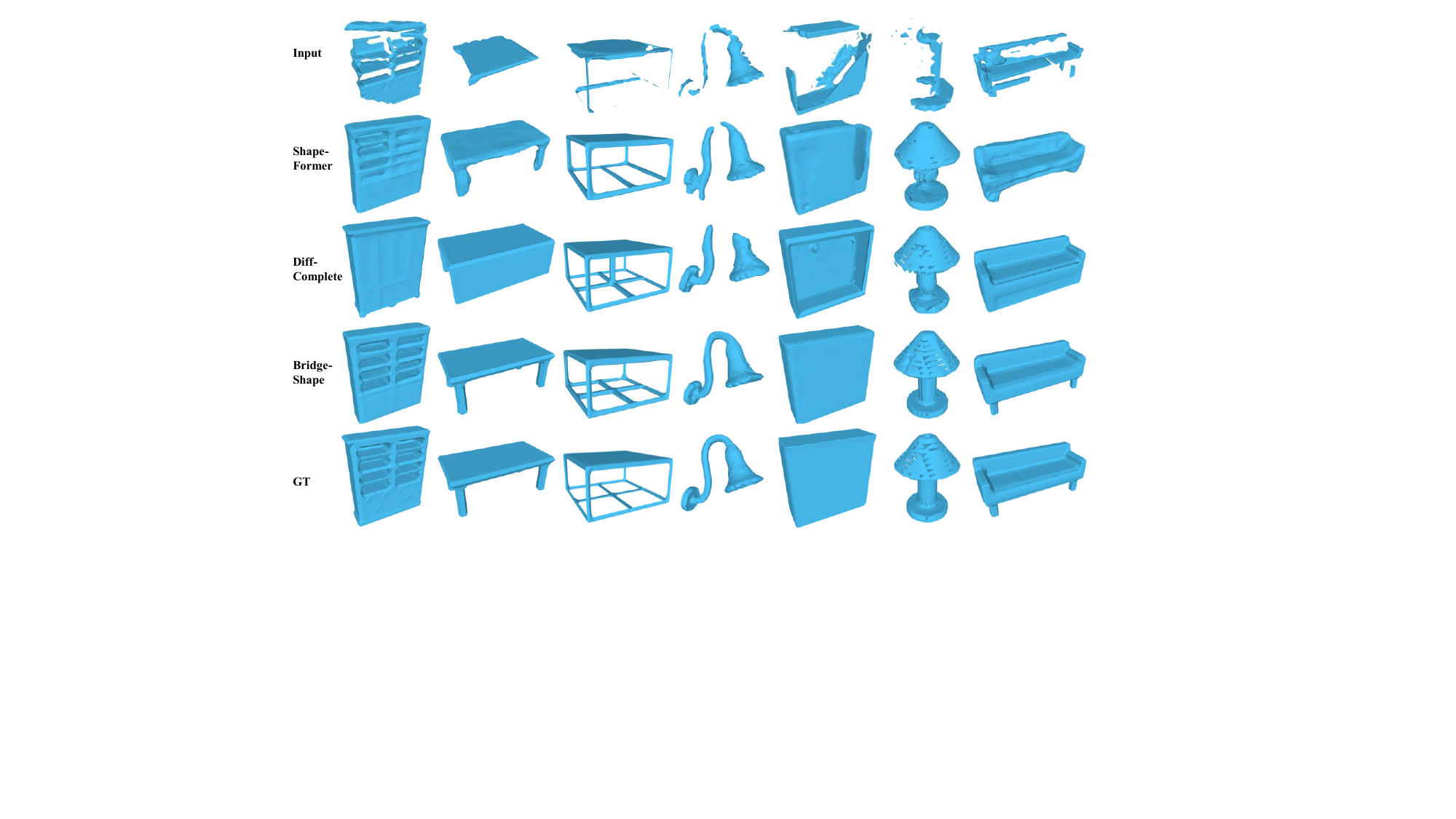}
\caption{Qualitative comparison of shape completion on 3D-EPN~\cite{dai2017shape}.}
\label{fig:seen}
\end{figure*}

\subsubsection{Noise Prediction and Inference.}
Our training objective is to accurately predict the noise injected at each timestep during the forward diffusion process. This is accomplished using a neural network \(\epsilon_{\theta}\) that estimates the noise \(\epsilon_{\theta}(\mathbf{z}_t, t)\) at every timestep \(t\). The training loss is defined as:
\begin{equation}
    \mathcal{L} = \|\epsilon_{\theta}(\mathbf{z}_t, t) - \frac{\mathbf{z}_t - \mathbf{z}_0}{\sigma_t}\|_2^2.
\end{equation}

During inference, we can run standard DDPM~\cite{ho2020denoising} to iteratively refine the complete shape’s latent representation, starting from \(\mathbf{z}_T\)(the latent code of the incomplete shape) and using reverse dynamics governed by the mean \(\mu_t(\mathbf{\hat{z}}_0, \mathbf{z}_T)\) and covariance \(\Sigma_t\) as:
\begin{equation}
p(\mathbf{z}_{t-1} \mid \mathbf{z}_t, \mathbf{\hat{z}_0}) = \mathcal{N}(\mathbf{z}_{t}; \mu_t(\mathbf{\hat{z}_0}, \mathbf{z}_T), \Sigma_t).
\end{equation}
This induces the same marginal density of Schrödinger bridge paths when \(\mathbf{\hat{z}}_0\) is closed to \(\mathbf{z}_0\)~\cite{i2sb}. 

\label{sec:method}

\begin{table*}[t!]
    \centering
    \setlength{\tabcolsep}{1mm}
     \begin{tabular*}{\textwidth}{l c c c c c c c c c c c c c c c c c c }
\toprule
 \textbf{}  & \multicolumn{2}{c}{Bag} & \multicolumn{2}{c}{Lamp} & \multicolumn{2}{c}{Bathtub} & \multicolumn{2}{c}{Bed} & \multicolumn{2}{c}{Basket} & \multicolumn{2}{c}{Printer} & \multicolumn{2}{c}{Laptop} & \multicolumn{2}{c}{Bench} & \multicolumn{2}{c}{Avg.}  \\
\cmidrule(lr){2-3} \cmidrule(lr){4-5} \cmidrule(lr){6-7} \cmidrule(lr){8-9} \cmidrule(lr){10-11} \cmidrule(lr){12-13} \cmidrule(lr){14-15} \cmidrule(lr){16-17} \cmidrule(lr){18-19}
\textbf{} & {CD~$\downarrow$} & {IoU~$\uparrow$} & {CD} & {IoU} & {CD} & {IoU} & {CD} & {IoU} & {CD} & {IoU} & {CD} & {IoU} & {CD} & {IoU} & {CD} & {IoU} & {CD} & {IoU} \\
\specialrule{0em}{2pt}{0pt}
\hline
\specialrule{0em}{1.5pt}{0pt}

3D-EPN & 5.01 & 73.8 & 8.07 & 47.2 & 4.21 & 57.9 & 5.84 & 58.4 & 7.90 & 54.0 & 5.15 & 73.6 & 3.90 & 62.0 & 4.54 & 48.3 & 5.58 & 59.4 \\
Few-Shot & 8.00 & 56.1 & 15.1 & 25.4 & 7.05 & 45.7 & 10.0 & 39.6 & 8.72 & 40.6 & 9.26 & 56.7 & 10.4 & 31.3 & 8.11 & 27.2 & 9.58 & 40.3 \\
IF-Nets & 4.77 & 69.8 & 5.70 & 50.8 &  4.72 & 55.0 & 5.34 & 60.7 & 4.44 & 50.2 & 5.83 & 70.5 & 6.47 & 58.3 & 5.03 & 49.7 & 5.29 & 58.1 \\
Auto-SDF & 5.81 & 56.3 & 6.57 & 39.1 & 5.17 & 41.0 & 6.01 & 44.6 & 6.70 & 39.8 & 7.52 & 49.9 & 4.81 & 51.1 & 4.31 & 39.5 & 5.86 & 45.2 \\
ConvONet & 5.10 & 70.8 & 5.42 & 52.6 & 4.96 & 60.4 & 5.42 & 63.2 & 6.16 & 54.6 & 5.56 & 72.1 & 4.78 & 57.3 & 4.69 & 49.6 & 5.26 & 60.1 \\
PatchComplete &  3.94 & 77.6 & \textbf{4.68} & 56.4 & 3.78 & 66.3 & 4.49 & 66.8 & 5.15 & 61.0 & 4.63 & 77.6 & 3.77 & 63.8 & 3.70 & 53.9 & 4.27 & 65.4 \\
DiffComplete & 3.86 & 78.3 & 4.80 & 57.9 & 3.52 & 68.9 & \textbf{4.16} & 67.1 & \textbf{4.94} & \textbf{65.5} & 4.40 & 76.8 & 3.52 & 67.4 & 3.56 & 58.2 & 4.10 & 67.5 \\
\specialrule{0em}{1.5pt}{0pt}
\hline
\specialrule{0em}{2pt}{0pt}
BridgeShape~\small{(Ours)}  & \textbf{3.70} & \textbf{80.0} & 4.94 & \textbf{62.7} & \textbf{3.41} & \textbf{69.8} & 4.32 & \textbf{69.6} & 5.50 & 65.1 & \textbf{4.09} & \textbf{81.4} & \textbf{3.13} & \textbf{72.7} & \textbf{3.38} & \textbf{59.1} & \textbf{4.06} & \textbf{70.1} \\
\bottomrule
\end{tabular*}
    \caption{Quantitative comparison with state-of-the-art methods~\cite {dai2017shape, wallace2019few, chibane2020implicit, mittal2022autosdf, peng2020convolutional, rao2022patchcomplete, chu2023diffcomplete} on synthetic objects~\cite{chang2015shapenet} of unseen categories. ( CD $\times 10^2$ and IoU $\times 10^2$)}
\label{tab:bridge_synthetic}
\end{table*}

\begin{table*}[t!]
\setlength{\tabcolsep}{2.1mm}
    \centering
     \begin{tabular*}{\textwidth}{l c c c c c c c c c c c c c c c c c c }
\toprule
 \textbf{}  & \multicolumn{2}{c}{Bag} & \multicolumn{2}{c}{Lamp} & \multicolumn{2}{c}{Bathtub} & \multicolumn{2}{c}{Bed} & \multicolumn{2}{c}{Basket} & \multicolumn{2}{c}{Printer} & \multicolumn{2}{c}{Avg.}  \\
\cmidrule(lr){2-3} \cmidrule(lr){4-5} \cmidrule(lr){6-7} \cmidrule(lr){8-9} \cmidrule(lr){10-11} \cmidrule(lr){12-13} \cmidrule(lr){14-15} 
\textbf{} & {CD~$\downarrow$} & {IoU~$\uparrow$} & {CD} & {IoU} & {CD} & {IoU} & {CD} & {IoU} & {CD} & {IoU} & {CD} & {IoU} & {CD} & {IoU}  \\
\specialrule{0em}{2pt}{0pt}
\hline
\specialrule{0em}{1.5pt}{0pt}

3D-EPN & 8.83 & 53.7&14.3 & 20.7&7.56 & 41.0&7.76 & 47.8&7.74 & 36.5&8.36 & 63.0&9.09 & 44.0 \\
Few-Shot & 9.10 & 44.9&11.9 & 19.6&7.77 & 38.2&9.07 & 34.9&8.02 & 34.3&8.30 & 62.2&9.02 & 38.6 \\
IF-Nets & 8.96 & 44.2&10.2 & 24.9&7.19 & 39.5&8.24 & 44.9&6.74 & 42.7&8.28 & 60.7&8.26 & 42.6 \\
Auto-SDF & 9.30 & 48.7&11.2 & 24.4&7.84 & 36.6&7.91 & 38.0&7.54 & 36.1&9.66 & 49.9&8.90 & 38.9 \\
ConvONet & 9.12 & 52.5&9.83 & 20.3&7.93 & 41.2&8.14 & 41.6&7.39 & 37.0&7.62 & 64.9&8.34 & 42.9 \\
PatchComplete &  8.23 & 58.3&9.42 & 28.4&6.77 & 48.0&7.24 & 48.4& 6.60 & 45.5&6.84 & 70.5&7.52 & 49.8 \\
DiffComplete & \textbf{7.05} & 48.5& \textbf{6.84} & 30.5&8.22 & 48.5&7.20 & 46.6&7.42 & \textbf{59.2} &\textbf{6.36} & \textbf{74.5}&7.18 & 51.3 \\
\specialrule{0em}{1.5pt}{0pt}
\hline
\specialrule{0em}{2pt}{0pt}
BridgeShape~\small{(Ours)}  & 7.67 & \textbf{61.2}  & 8.07 & \textbf{36.3} & \textbf{6.28} & \textbf{50.9} & \textbf{6.87} & \textbf{51.3}  & \textbf{6.20} & 50.2  & 6.83 & 71.0 & \textbf{6.99} & \textbf{53.5} \\

\bottomrule
\end{tabular*}
    \caption{Quantitative comparison with state-of-the-art methods~\cite {dai2017shape, wallace2019few, chibane2020implicit, mittal2022autosdf, peng2020convolutional, rao2022patchcomplete, chu2023diffcomplete} on real-world objects~\cite{dai2017scannet} of unseen categories. ( CD $\times 10^2$ and IoU $\times 10^2$)}

\label{tab:bridge_real}
\end{table*}

\section{Experiments}

\subsection{Experimental Settings}

\subsubsection{Datasets.} We evaluate on two large-scale shape completion benchmarks. 3D‑EPN~\cite{dai2017shape} comprises 25,590 training and 5,384 testing instances across eight ShapeNet categories, each with six partial scans (\(32^3\) TSDF) and corresponding complete shapes (\(32^3\), \(64^3\), or \(128^3\) TUDF). 
~
PatchComplete~\cite{rao2022patchcomplete} includes synthetic ShapeNet data and real-world scan data from ScanNet \cite{dai2017scannet}. For ShapeNet, 18 categories are used for training and 8 for testing, with 3,202 training models and 1,325 test models, each having four partial scans. The ScanNet data consists of objects extracted from bounding boxes, with complete shapes provided by Scan2CAD \cite{avetisyan2019scan2cad}. All objects are represented as \( 32^3 \) TSDF units. These datasets allow us to evaluate our method on both synthetic and real-world data, testing its ability to handle unseen categories.

\subsubsection{Evaluation.}
We evaluate our method using standard metrics for shape completion. On the 3D-EPN dataset~\cite{dai2017shape}, we report the mean \(l_1\) error across all voxels of the TUDF predictions. For PatchComplete benchmark~\cite{rao2022patchcomplete}, we use the \(l_1\) Chamfer Distance (CD) and Intersection over Union (IoU) to evaluate the geometry of predicted shapes. 10K points are sampled on surfaces for CD calculation.

\subsection{Evaluation on known object categories}
The quantitative and qualitative comparison with state-of-the-art methods~\cite{dai2017shape, zhang2022probabilistic, mittal2022autosdf, zheng2022sdf, peng2020convolutional, rao2022patchcomplete, chu2023diffcomplete, yan2022shapeformer} on the 3D-EPN dataset~\cite{dai2017shape} is presented in Table~\ref{tab:epn_bench} and Figure~\ref{fig:seen}.
For the probabilistic methods, we report the average results across five inferences with different initializations. 
In comparison to the second-ranked method DiffComplete~\cite{chu2023diffcomplete}, BridgeShape reduces the l1 error by approximately 26\% (from 0.053 to 0.039), while producing high-fidelity, realistic shapes with fewer surface artifacts.
%
This is because DiffComplete reverses from unstructured Gaussian noise and relies on deep feature interactions to inject incomplete shape information, while our approach directly starts the generation process from the input incomplete shapes, providing a far more informative prior. Additionally, compared to GAN-based SDF-StyleGAN~\cite{zheng2022sdf}, autoregressive methods AutoSDF~\cite{mittal2022autosdf} and ShapeFormer~\cite{yan2022shapeformer}, our diffusion bridge-based generative model exhibits enhanced mode coverage and superior sampling quality. 
Moreover, while deterministic approaches~\cite{dai2017shape, peng2020convolutional, rao2022patchcomplete} perform one-step mappings, BridgeShape refines shapes iteratively during the diffusion bridge process, significantly enhancing accuracy. 

\subsection{Evaluation on Unseen Object Categories}
\begin{figure}[!t]
\centering
\includegraphics[width=0.48\textwidth]{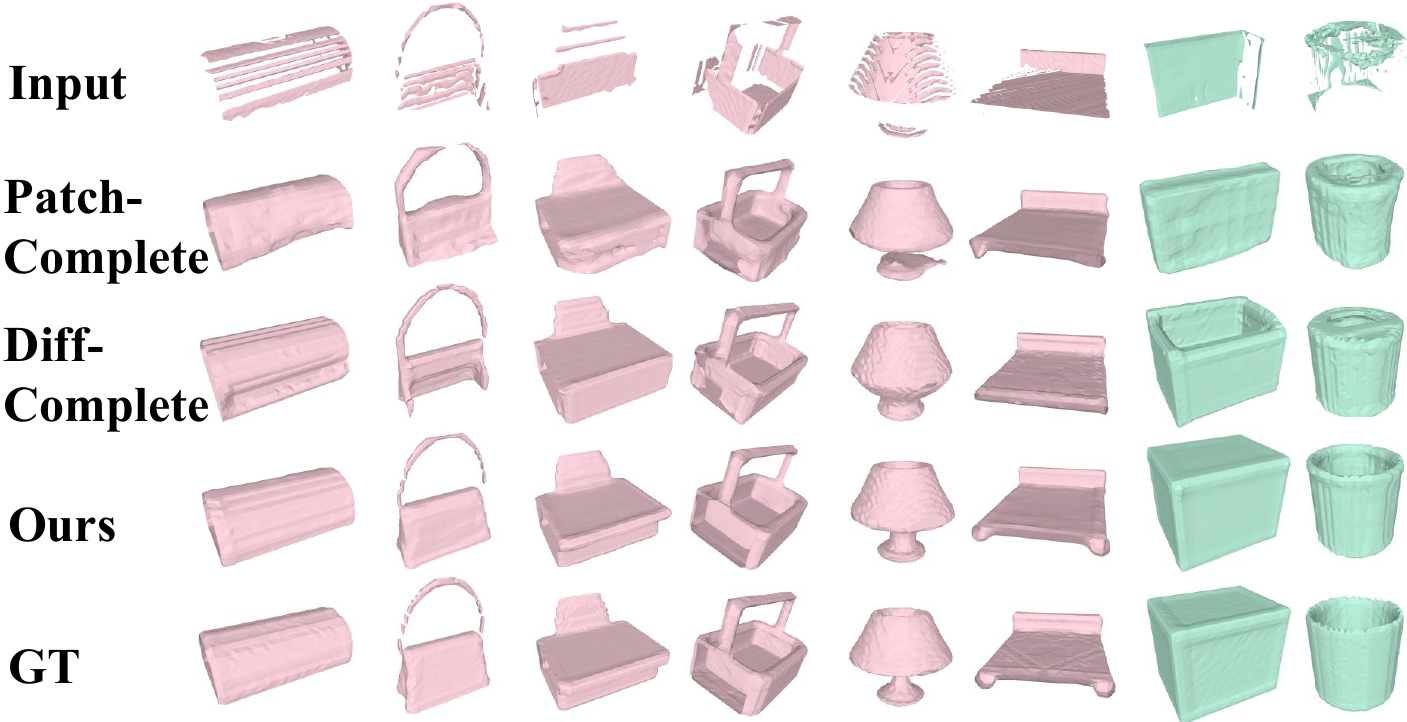}
\caption{Qualitative comparison on the synthetic (pink) ShapeNet~\cite{chang2015shapenet} dataset and real-world (green) ScanNet~\cite{dai2017scannet} dataset.}
\label{fig:unseen}
\end{figure}

We evaluate BridgeShape’s ability to generalize by comparing against state-of-the-art methods—3D-EPN~\cite{dai2017shape}, Few-Shot~\cite{wallace2019few}, IF-Nets~\cite{chibane2020implicit}, Auto-SDF~\cite{mittal2022autosdf}, ConvONet~\cite{peng2020convolutional}, PatchComplete~\cite{rao2022patchcomplete}, DiffComplete~\cite{chu2023diffcomplete}—on synthetic ShapeNet~\cite{chang2015shapenet} and real-world ScanNet data~\cite{dai2017scannet}. As shown in Table~\ref{tab:bridge_synthetic} and Table~\ref{tab:bridge_real}, BridgeShape excels in handling unseen categories, outperforming them in terms of CD and IoU on average. This can be attributed to the fact that BridgeShape enforces globally coherent transport via the Schrödinger bridge formulation, effectively modeling shared structures among different categories. This leads to superior performance on unseen categories, including real-world scans that are typically cluttered and noisy. 
Figure~\ref{fig:unseen} provides compelling visual evidence that BridgeShape robustly generalizes to unseen categories, producing high-fidelity completions that preserve fine geometric details on both synthetic and real-world datasets.

\subsection{Ablation Studies}
\label{sec:ablation}
We perform component ablations on 3D-EPN~\cite{dai2017shape}, reporting average performance across all categories unless otherwise noted.  

\begin{table}[!t]
\centering
\setlength{\tabcolsep}{3mm}
\begin{tabular}{l c}
\toprule 
Mechanisms  & $l_1$-err.~$\downarrow$\\  
\specialrule{0em}{2pt}{0pt}
\hline
\specialrule{0em}{1.5pt}{0pt}
Conventional diffusion & 0.047 \\
DSB & 0.045 \\
DSB + Stochasticity~\small{(Ours)}  & \textbf{0.039} \\
\bottomrule
\end{tabular}
\caption{Ablation on different diffusion mechanisms.}
\label{tab:mechanisms}
\end{table}

\subsubsection{Effects of DSB.}
One of the key innovations in BridgeShape is the DSB, which explicitly models optimal transport between the latent distributions of incomplete and complete shapes. To assess its impact, we replace the DSB with a conventional diffusion paradigm with an additional conditional branch, similar to DiffComplete~\cite{chu2023diffcomplete}. However, as shown in Table~\ref{tab:mechanisms}, this approach yields suboptimal results, underscoring the importance of explicitly modeling the transport process to achieve superior shape completions. Moreover, injecting stochasticity into the latent distribution of incomplete shapes can further boost performance.

\subsubsection{Effects of Depth-Enhanced VQ-VAE.}
The Depth-Enhanced VQ-VAE serves as the foundation for the DSB. Unlike a standard VQ-VAE~\cite{razavi2019generating,van2017neural,cheng2023sdfusion}, which captures a compact latent representation but struggles to retain structural details, our approach integrates self-projected multi-view depth information enriched with DINOv2~\cite{oquab2023dinov2} features to enhance geometric perception. To assess its effectiveness, we compare it against a standard VQ-VAE without depth features. 
~
As shown in Table~\ref{tab:VQVAE}, integrating multi-view depth into our VQ-VAE halves its reconstruction $l_1$ error (0.0042→0.0021), yields a 0.002 reduction in completion $l_1$ error, and boosts IoU by $\sim$2\% (92.36\%→94.41\%). Removing depth reverses these gains, underscoring the critical role of multi-view depth. Additional ablation studies and qualitative results are provided in the supplementary material.

\begin{table}[t!]
\centering
\setlength{\tabcolsep}{1.7mm}
 \begin{tabular*}{0.99\linewidth}{l c c c}
\toprule
Variants & Rec. $l_1$~$\downarrow$ & Comp. $l_1$~$\downarrow$ & Comp. IoU~$\uparrow$ \\  
\specialrule{0em}{2pt}{0pt}
\hline
\specialrule{0em}{1.5pt}{0pt}
W/o. Depth & 0.0042 & 0.0413 & 92.36\% \\ 
W. Depth (Ours) & \textbf{0.0021} & \textbf{0.0389} & \textbf{94.41\%}\\ 
\bottomrule
\end{tabular*}
\caption{Ablation on the effect of Depth.}
\label{tab:VQVAE}
\end{table}

\subsection{Results on Higher Voxel Resolution}
While our primary experiments use a voxel resolution of $32^3$, we further evaluate on 3D-EPN~\cite{dai2017shape} at $64^3$ and $128^3$ to assess scalability (Table~\ref{tab:resolution}). Higher resolutions enable finer-grained shape completion and yield increasingly higher average accuracy. These results confirm that BridgeShape effectively leverages increased resolution to enhance completion quality. Qualitative results are provided in the supplementary material.

\begin{table}[!t]
\centering
\setlength{\tabcolsep}{2mm}
 \begin{tabular*}{0.98\linewidth}{l  c c c}
\toprule 

Categories & $l_1$-err. ($32^3$) & $l_1$-err. ($64^3$) & $l_1$-err. ($128^3$)\\
\specialrule{0em}{2pt}{0pt}
\hline
\specialrule{0em}{1.5pt}{0pt}
Chair & 0.055 & 0.045 & \textbf{0.044} \\
Table & 0.059 & \textbf{0.053} & 0.054\\
Sofa & \textbf{0.047} & 0.051 & 0.050\\
Lamp & 0.038 & 0.034 & \textbf{0.026} \\
Plane & 0.012 & \textbf{0.009} & \textbf{0.009} \\
Car & \textbf{0.023} & 0.024 & 0.024\\
Dresser & \textbf{0.055} & 0.062 & 0.058\\ 
Watercraft & 0.022 & \textbf{0.019} & \textbf{0.019}\\
Avg. & 0.039 & 0.037 & \textbf{0.036} \\

\bottomrule
\end{tabular*}
\caption{Quantitative results on 3D-EPN. Low is better.}
\label{tab:resolution}
\end{table}

\label{sec:experiments}

\section{Conclusion}
We present BridgeShape, a latent diffusion Schrödinger bridge framework for 3D shape completion. By enriching a compact latent space with multi-view depth features, BridgeShape efficiently transports incomplete shapes to complete ones, achieving high-fidelity and structurally consistent results. Experiments on large-scale benchmarks show superior performance, fine geometric preservation, and strong generalization to unseen categories.

\section{Acknowledgments}
This work was supported by the National Natural Science Foundation of China (No. T2322012, No. 62572240).

\bibliography{aaai2026}

\clearpage

\begin{center}
    \LARGE \textbf{Appendix}
\end{center}

\renewcommand{\thesection}{\Alph{section}}

In this supplementary material, we provide additional details to complement the main manuscript.  Specifically, we begin by describing the implementation details, including a in-depth overview of the model architecture and the process of multi-view depth maps. Next, we present a series of additional experiments to further analyze our method. Furthermore, we offer both quantitative and qualitative results to provide a more comprehensive understanding of its performance. Finally, we discuss several failure cases and the limitations of our approach.

\setcounter{section}{0}
\section{Implementation Details}


\begin{figure}[!ht]
\centering
\includegraphics[width=0.47\textwidth]{ 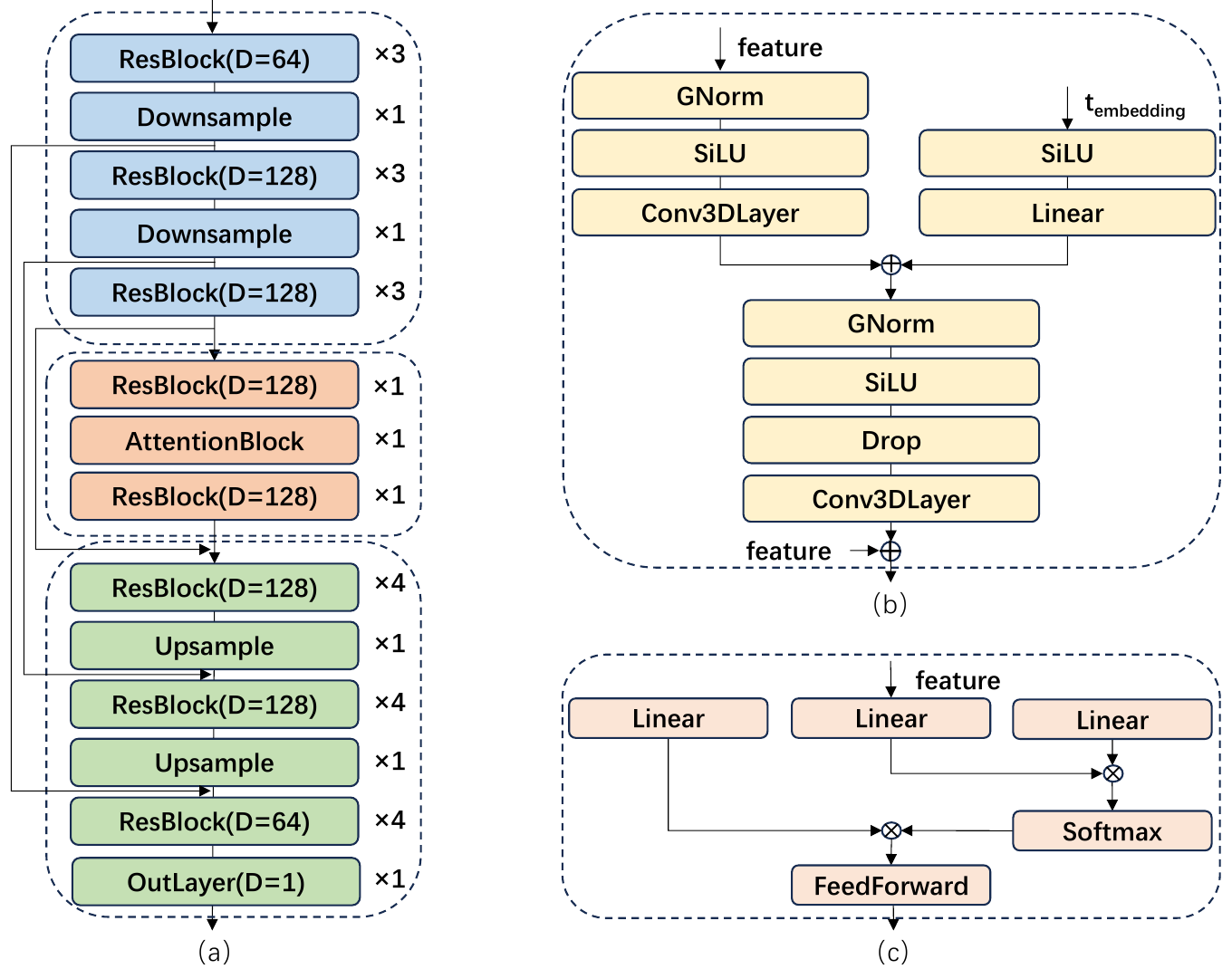}
\caption{(a) Architecture of the diffusion model, consisting of an encoder, intermediate blocks, and a decoder. (b) Detailed structure of the ResBlock. (c) Detailed structure of the AttentionBlock.}
\label{fig:architecture}
\end{figure}

\subsection{Model Architecture Details}

Figure~\ref{fig:architecture} (a) presents the detailed architecture of our diffusion model, which consists of an encoder, intermediate blocks, and a decoder. The encoder is composed of multiple ResBlock layers and downsampling layers, which reduce the resolution by half at each step. The intermediate block combines two ResBlocks with an AttentionBlock for feature refinement. The decoder reconstructs the spatial dimensions using a mirrored architecture, employing upsampling layers (with a 2× resolution scaling) in conjunction with ResBlocks. Skip connections are used to establish direct pathways between corresponding encoder and decoder stages, ensuring effective information flow. Figure~\ref{fig:architecture} (b) illustrates the detailed structure of the ResBlock, which receives features from the previous layer and a time embedding as inputs. Figure~\ref{fig:architecture} (c) shows the detailed structure of the AttentionBlock.

\subsection{Experiment and training settings} We begin by converting the TUDF data into meshes using the Marching Cubes algorithm~\cite{lorensen1987marching}. Next, using Blender, each mesh is rendered from three canonical viewpoints (front, top, and left) to produce three depth maps. Multi-view rendering from canonical space ensures consistent alignment while mitigating distortion, scale bias, and occlusion. These depth maps serve as input for training our Depth-Enhanced VQ-VAE model on complete 3D shapes for 400k steps. During this stage, the DINOv2~\cite{oquab2023dinov2} model remains frozen, leveraging its pretrained weights for feature extraction. The VQ-VAE is trained with a two-stage procedure: Stage I removes the cross-attention layer and train the model solely on complete shapes without depth information to learn a stable base latent representation; 
Stage II reintroduces the cross-attention layer and fuses multi-view self-projected depth features to enrich the latent space, improving the model’s ability to capture fine geometric details from multiple perspectives. We distill these 2D priors into the latent representation only during training, unlike SDF-Diffusion~\cite{shim2023diffusion}. Next, we train our diffusion bridge model along with an additional encoder for another 200k steps.
~
For optimization, we use the Adam optimizer~\cite{diederik2014adam} with a learning rate of \(1\times10^{-4}\) and batch size 16 to train the Depth-Enhanced VQ-VAE, whose latent code \(z\in\mathbb{R}^{16^3\times3}\) employs a codebook of 8192 entries. For the latent diffusion Schrödinger bridge, we follow prior works~\cite{i2sb}—drift \(f:=0\) (the learned score implicitly restores it, ensuring optimal-transport validity), symmetric noise schedule over \(T=1000\) steps—and adopt AdamW~\cite{loshchilov2017decoupled} at \(2\times10^{-4}\) with batch size 24. We inject Gaussian noise into the latent distribution of incomplete shapes before constructing the optimal transport process at a scale of 1×.
~
All experiments were conducted on Ubuntu 20.04 with PyTorch 2.0 (CUDA 11.8) using a single NVIDIA RTX 4090 GPU (24 GB VRAM).
~
For the 3D-EPN benchmark~\cite{dai2017shape}, we train category-specific models, while on PatchComplete benchmark~\cite{rao2022patchcomplete}, a unified model is trained across all categories to promote generalization. During inference, we set 3 denoising steps for the diffusion process. 

\begin{table}[t]
\centering
\setlength{\tabcolsep}{5pt}
\begin{tabular}{lccccc}
\toprule
    Variant & \shortstack{VQ-VAE} & DSB & \shortstack{Conv.Diff.} & $64^3$ & $128^3$ \\
    \midrule
    (a) &            & \checkmark &            & 0.047 & 0.045 \\
    (b) & \checkmark &            & \checkmark & 0.047 & 0.046 \\
    (c) & \checkmark & \checkmark &            & \textbf{0.037} & \textbf{0.036} \\
    (d) &            &            & \checkmark & 0.052 & -- \\
    \bottomrule
  \end{tabular}
\caption{Component ablation: $l_1$ error at $64^3$ and $128^3$.}
\label{tab:main_ablation}
\end{table}

\begin{table}[t]
  \centering
  \small
  \setlength{\tabcolsep}{2pt}
  \begin{tabular}{lcccc}
    \toprule
    Method       & Steps~($\downarrow$) & GFLOPs/step~($\downarrow$) &  Runtime (s)~($\downarrow$) \\
    \midrule
    DiffComplete & 100   & 159.5   & 3.2  \\
    BridgeShape (Ours) & \textbf{3} & \textbf{33.6}  & \textbf{0.04} \\
    \bottomrule
  \end{tabular}
  \caption{Inference cost (GFLOPs) and runtime per sample.}
  \label{tab:inference_efficiency}
\end{table}

\begin{table}[t]
\centering
\begin{tabular}{lcc}
\toprule
Strategies & $l_1$-err.~($\downarrow$) & Memory (GB)~($\downarrow$) \\
\midrule
Concatenation      & \textbf{0.0553}    & 19.60               \\
Max Pooling        & 0.0558    & \textbf{16.28}               \\
Average Pooling~\small{(Ours)}       & 0.0554   & \textbf{16.28} \\
\bottomrule
\end{tabular}
\caption{Ablation study on the effect of different feature aggregation strategies. (batch size 12)}
\label{tab:Aggregation}
\end{table}

\begin{figure*}[!t]
\centering
\includegraphics[width=0.97\textwidth]{ 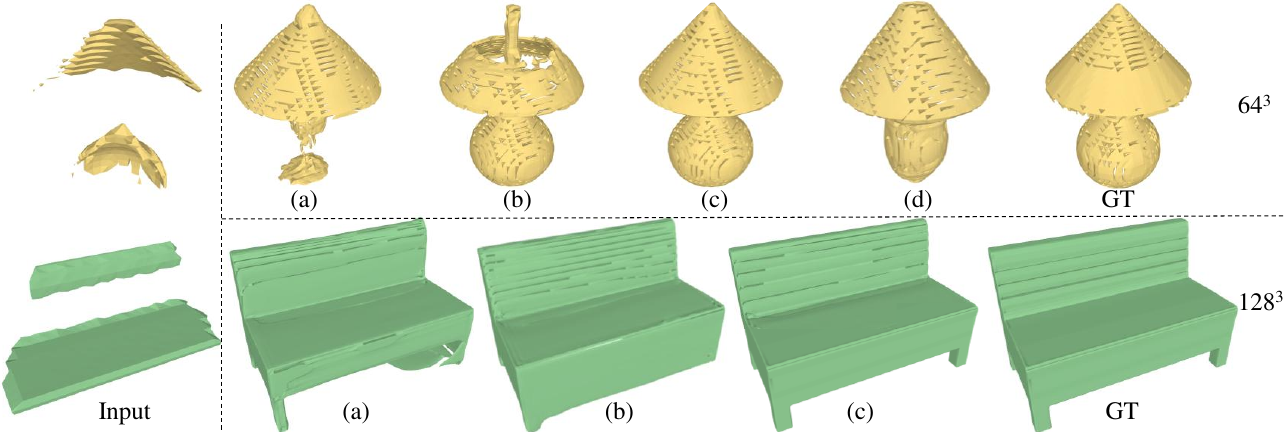}
\caption{Visuals: Component ablation.}
\label{fig:abla22}
\end{figure*}

\begin{figure*}[!t]
\centering
\includegraphics[width=0.97\textwidth]{ 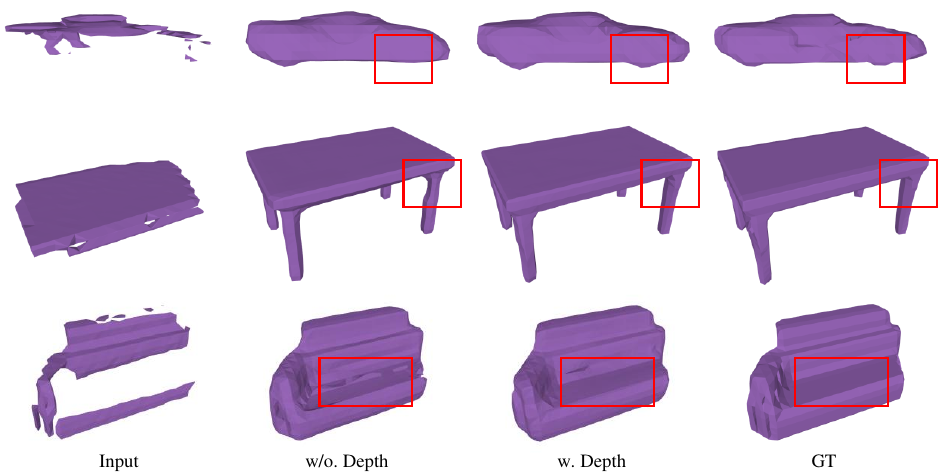}
\caption{Effect of depth on completion performance.}
\label{fig:depth}
\end{figure*}

\section{Additional Experiments}

\subsection{Component Ablation Study}

To assess the individual contributions of each module in BridgeShape, we conduct a series of ablation experiments on 3D-EPN~\cite{dai2017shape} at voxel resolutions of $64^3$ and $128^3$. As summarized in Table~\ref{tab:main_ablation}, we evaluate the following variants:
\begin{itemize}
    \item[(a)] \textbf{Voxel-space diffusion Schrödinger(DSB):} Apply the diffusion Schrödinger bridge directly in the original voxel grid, without latent compression.
    \item[(b)] \textbf{Depth-Enhanced VQ-VAE + standard diffusion:} Use our Depth-Enhanced VQ-VAE to encode shapes into latent space, then perform conventional conditional diffusion without DSB.
    \item[(c)] \textbf{BridgeShape (ours):} Combine the Depth-Enhanced VQ-VAE with DSB.
    \item[(d)] \textbf{Voxel-space standard diffusion:} A baseline using conventional diffusion directly in voxel space ( DiffComplete~\cite{chu2023diffcomplete}).
\end{itemize}
Combining the Depth-Enhanced VQ-VAE with standard diffusion (b) reduces the $l_1$ error from 0.052 to 0.047, highlighting the benefit of a compact yet structurally informative latent space. Using DSB in voxel space (a) yields a similar $l_1$ error of 0.047. Our full model (c) further lowers the error to 0.037 at $64^3$ and 0.036 at $128^3$, demonstrating the synergistic benefit of combining both components. Figure~\ref{fig:abla22} shows side-by-side qualitative comparisons.

\subsection{Inference Cost and Runtime Comparison}
In addition to superior completion accuracy, BridgeShape offers a dramatic improvement in inference efficiency (see Table~\ref{tab:inference_efficiency}). BridgeShape requires only three reverse-diffusion steps rather than the hundreds typical of standard noise-to-data models. At $32^3$ input resolution and batch size 1,, each inference step costs approximately 33.6 GFLOPs, for a total of $\sim$100.8 GFLOPs, and the full inference process (encoding, 3 reverse steps, and decoding) completes in just 0.04~s. By contrast, DiffComplete~\cite{chu2023diffcomplete} performs 100 diffusion steps at 159.5 GFLOPs each ($\sim$15 950 GFLOPs total), requiring 3.2~s per sample—over 80× slower.

\subsection{Effects of View Aggregation Mechanism} To assess the influence of different multi-view depth feature aggregation strategies in VQ-VAE, we conduct an ablation study on 3D-EPN~\cite{dai2017shape} chair class, with the results summarized in Table~\ref{tab:Aggregation}. We evaluate three feature aggregation strategies along the view dimension: average pooling, concatenation, and max pooling. The study investigates both the final completion performance and GPU memory consumption during VQ-VAE training, with GPU memory usage reported for a batch size of 12. 
For cross-attention, 3D features are flattened into 1D tokens for alignment.
Our results show that average pooling achieves a comparable $l_1$-error to concatenation, while significantly reducing memory consumption. In contrast, max pooling leads to a slight drop in accuracy, likely due to its tendency to retain only the most dominant responses, discarding fine-grained geometric details. These findings suggest that average pooling effectively captures complementary information across views while maintaining computational efficiency, making it a well-balanced choice for feature aggregation.

\begin{table}[t!]
\centering
\begin{tabular}{cc}
\toprule
Variants & $l_1$-err.~($\downarrow$) \\
\midrule
0 Views & 0.0563 \\
1 View & 0.0558 \\
3 Views (Ours) & \textbf{0.0554} \\
6 Views & \textbf{0.0554} \\
\bottomrule
\end{tabular}
\caption{Effect of different numbers of depth views.}
\label{tab:view}
\end{table}

\subsection{Ablation on the number of projections}
Table~\ref{tab:view} provides a quantitative evaluation of the effect of incorporating different numbers of depth views (0, 1, 3, and 6 views) on the 3D-EPN~\cite{dai2017shape} chair class. The results show that increasing the number of depth views consistently improves completion accuracy. Notably, the performance plateaus after 3 views, indicating that additional views beyond this threshold do not yield significant improvements. Based on this observation, we select 3 views for the experimental configuration, as it strikes an optimal balance between enhanced accuracy and computational efficiency.

\subsection{Ablation on the Scale of Stochasticity}
We evaluate the impact of Gaussian noise scale injected into the latent distribution of incomplete shapes on final completion performance on 3D-EPN~\cite{dai2017shape}. As shown in Table~\ref{tab:noise_scale}, a scale of $1\times$ achieves the lowest $l_1$ error, confirming that this level of stochasticity most effectively stabilizes the optimal transport process and compensates for missing geometry.

\begin{table}[t!]
\centering
\begin{tabular}{cc}
\toprule
Noise Scale & $l_1$-err.~($\downarrow$) \\
\midrule
0.5×        & 0.043            \\
\textbf{1×} (Ours)  & \textbf{0.039}   \\
2×          & 0.044            \\
\bottomrule
\end{tabular}
\caption{Effect of noise scale.}
\label{tab:noise_scale}
\end{table}

\begin{figure*}[!t]
\centering
\includegraphics[width=0.97\textwidth]{ 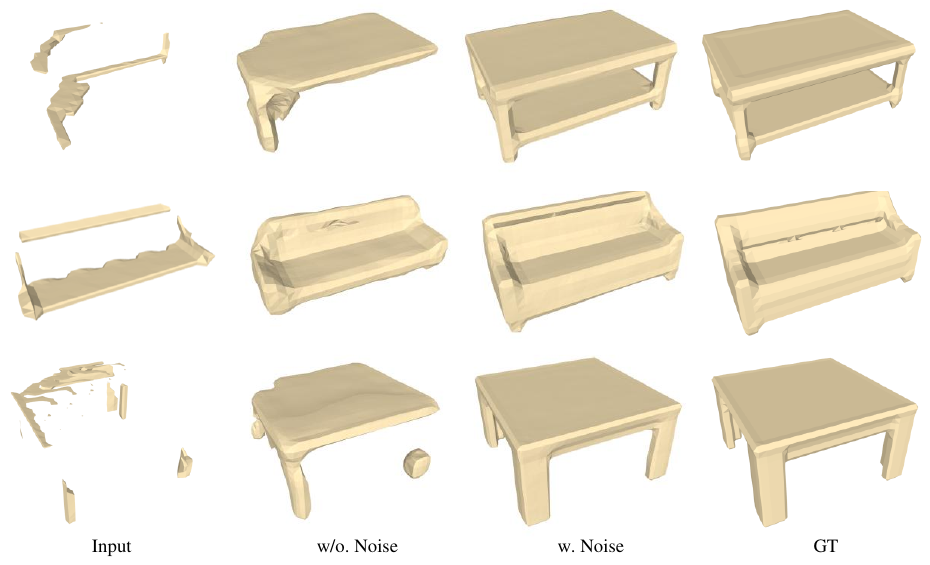}
\caption{Effect of stochasticity on completion performance.}
\label{fig:noise}
\end{figure*}

\begin{table}[!t]
\centering
\begin{tabular}{ccc}
\toprule
        Methods  & CD~($\downarrow$)  &F1~($\uparrow$)  \\
        \midrule[0.3pt]
        AdaPoinTr~\cite{yu2022adapointr}   &  21.52 & 0.286 \\
        BridgeShape (Ours) & \textbf{15.60}  & \textbf{0.518} \\
\bottomrule
\end{tabular}
\caption{Comparison with point cloud completion approach. ({$\displaystyle \ell ^{1}$} CD $\times 10^3$ and F1-Score@1\%)}
\label{tab:PCC}
\end{table}

\begin{table}[t]
\centering
\footnotesize
\setlength{\tabcolsep}{2.6pt}  
\begin{tabular}{lcccc}
\toprule
Size & Rec.\,$l_1\downarrow$ & Comp.\,$l_1\downarrow$ & Comp. IoU$\uparrow$ &  Time (norm.)$\downarrow$ \\
\midrule
4096   & 0.0034 & 0.040 & 92.23\% & \textbf{1.00×} \\
8192 (Ours)     & \textbf{0.0021} & \textbf{0.039} & \textbf{94.41}\% & 1.07×        \\
16384            & 0.0022 & \textbf{0.039} & 94.37\% & 1.12×        \\
\bottomrule
\end{tabular}
\caption{Effect of codebook size.}
\label{tab:codebook}
\end{table}

\subsection{Comparison with Point Cloud Completion Approach}
We compare our method with the state-of-the-art AdaPoinTr~\cite{yu2022adapointr} in the task of point cloud completion on 3D-EPN~\cite{dai2017shape} chair class. For fair comparison, we adopt the experimental settings commonly used by AdaPoinTr, where 2048 points are sampled from the incomplete shape surface and 8192 points from the complete shape surface for training AdaPoinTr. To ensure consistency in evaluation, we sample 8192 points from our predicted shape. We report the $\displaystyle \ell ^{1}$ version of Chamfer Distance (CD) and F1-Score as evaluation metrics. As shown in Table~\ref{tab:PCC}, our method achieves significantly improved completion performance. This result underscores the effectiveness of our approach.

\subsection{Ablation on the Size of the Codebook}
To evaluate the impact of codebook capacity on representation quality and completion cost, we compare three sizes (Table~\ref{tab:codebook}). While larger sizes offer only marginal gains in VQ-VAE's reconstruction fidelity and final completion accuracy, they incur higher inference cost—making 8192 the best trade-off.

\begin{figure*}[!t]
\centering
\includegraphics[width=0.97\textwidth]{ 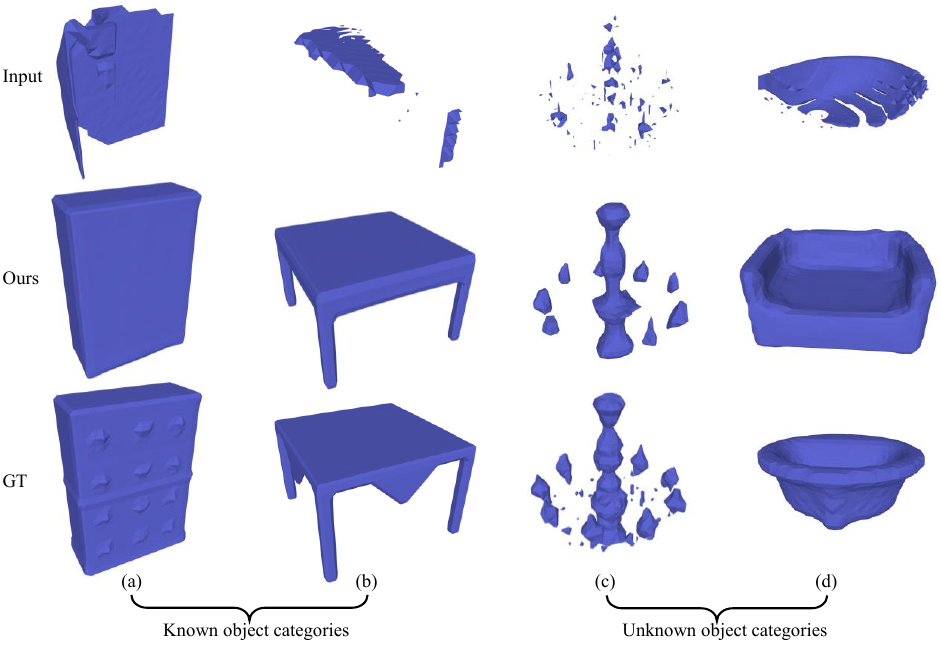}
\caption{Example of failure cases.}
\label{fig:failure}
\end{figure*}

\section{More Quantitative Visualizations}
\subsection{Effect of Depth on Completion Performance}
Figure~\ref{fig:depth} provides qualitative results of the impact of incorporating self-projected multi-view depth information on shape completion results. The figure demonstrates that using multi-view depth maps improves geometric consistency and the recovery of fine-grained details—for example, for example, in the completion of car tire, chair leg, and sofa surface.

\subsection{Effect of Stochasticity on Completion Performance} Figure~\ref{fig:noise} presents qualitative results illustrating the impact of injecting stochasticity into the latent distribution of incomplete shapes before constructing the optimal transport process. The figure highlights that this design is particularly effective in handling extremely sparse incomplete shapes, where significant uncertainty in the missing regions can hinder the establishment of a robust optimal transport path, thereby reducing completion accuracy.

\begin{figure*}[!t]
\centering
\includegraphics[width=0.97\textwidth]{ 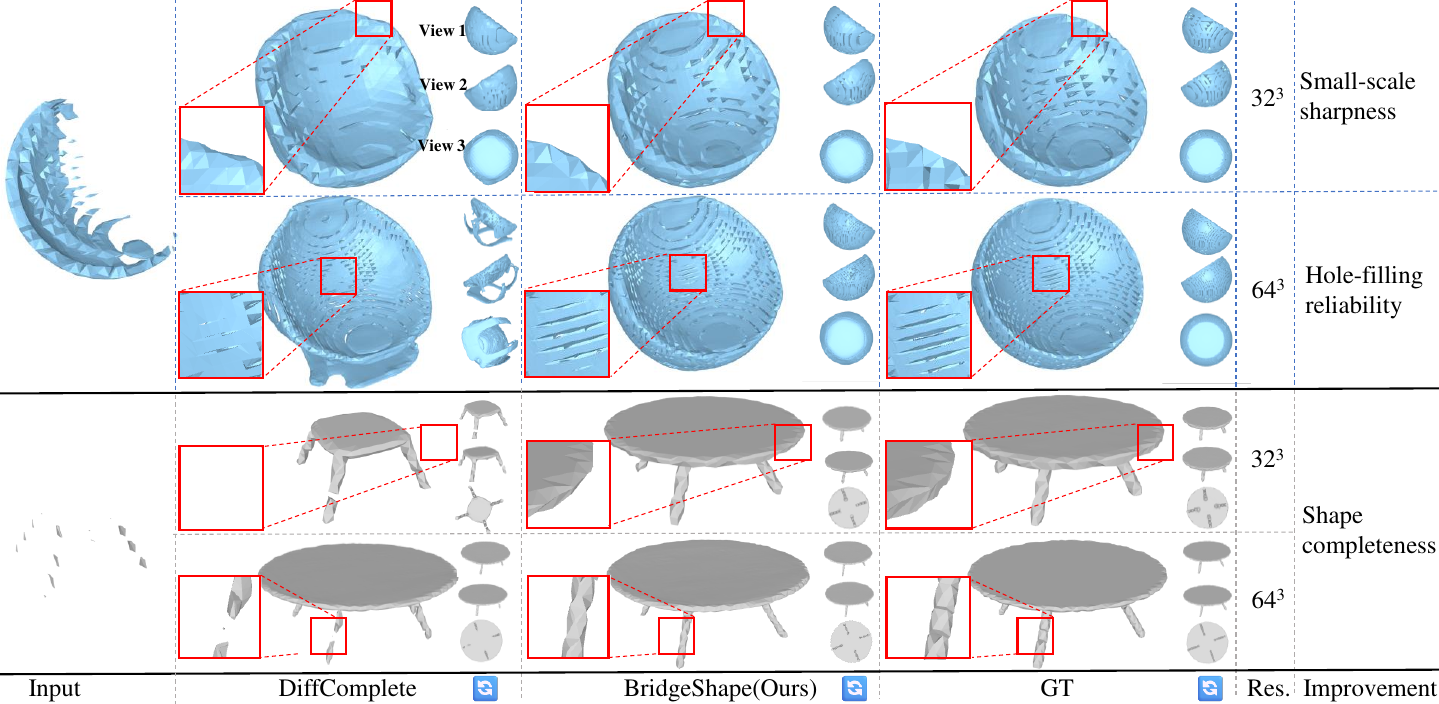}
\caption{BridgeShape vs.\ DiffComplete~\cite{chu2023diffcomplete} at $32^3$ and $64^3$.}
\label{fig:abla1}
\end{figure*}

\begin{figure*}[!t]
\centering
\includegraphics[width=0.97\textwidth]{ 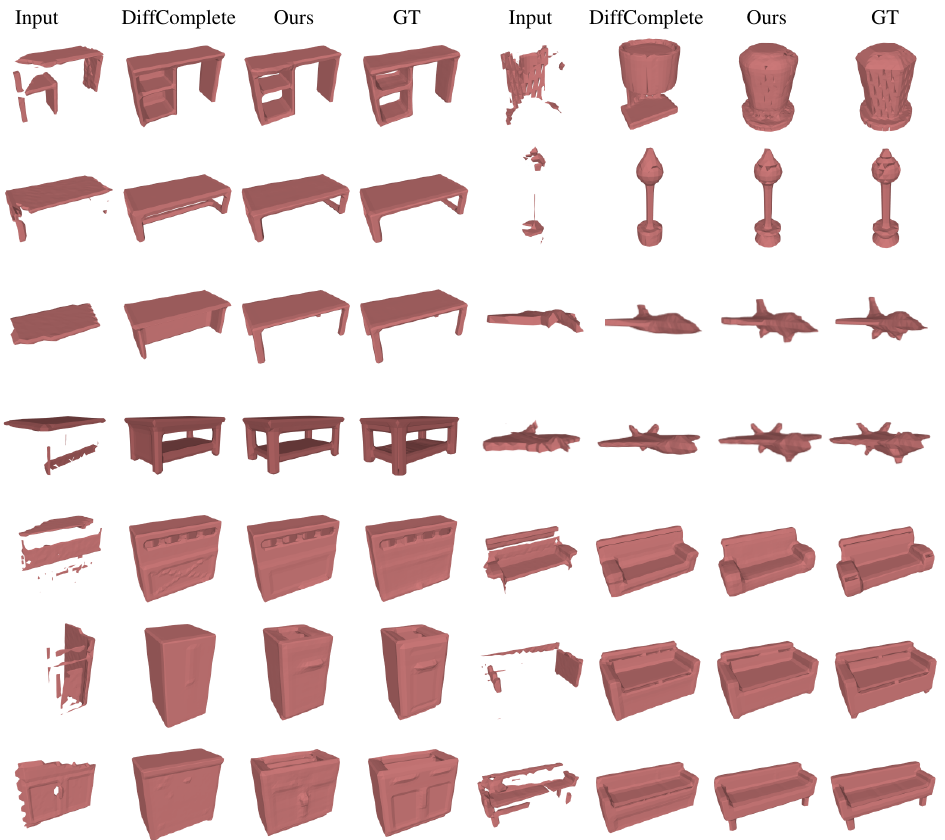}
\caption{Visual comparison with DiffComplete~\cite{chu2023diffcomplete} on 3D-EPN~\cite{dai2017shape}.}
\label{fig:EPN_more}
\end{figure*}

\subsection{Fine-Grained Visualization Comparisons}

Figure~\ref{fig:abla1} compares BridgeShape against DiffComplete~\cite{chu2023diffcomplete} on 3D-EPN~\cite{dai2017shape} at both $32^3$ and $64^3$ resolutions. BridgeShape delivers more accurate hole filling, sharper recovery of fine geometry, and stronger global completeness—benefits that are even more pronounced at the higher $64^3$ resolution.

\subsection{Visualizations on Known Object Categories}
Figure~\ref{fig:EPN_more} displays qualitative results comparing the state-of-the-art DiffComplete~\cite{chu2023diffcomplete} with our BridgeShape across different known object categories. As shown, our method produces shape completions that are more realistic and visually coherent. Furthermore, our completions are highly accurate, closely aligning with the ground truth shapes.

\begin{figure*}[!t]
\centering
\includegraphics[width=0.97\textwidth]{ 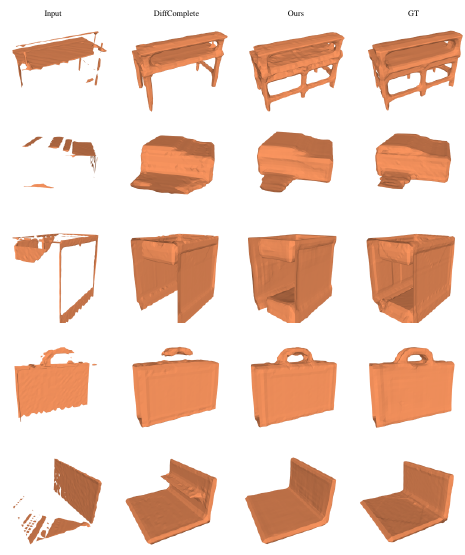}
\caption{Visual comparison with DiffComplete~\cite{chu2023diffcomplete} on Shapenet~\cite{chang2015shapenet}.}
\label{fig:shapenet_more}
\end{figure*}

\begin{figure*}[!t]
\centering
\includegraphics[width=0.97\textwidth]{ 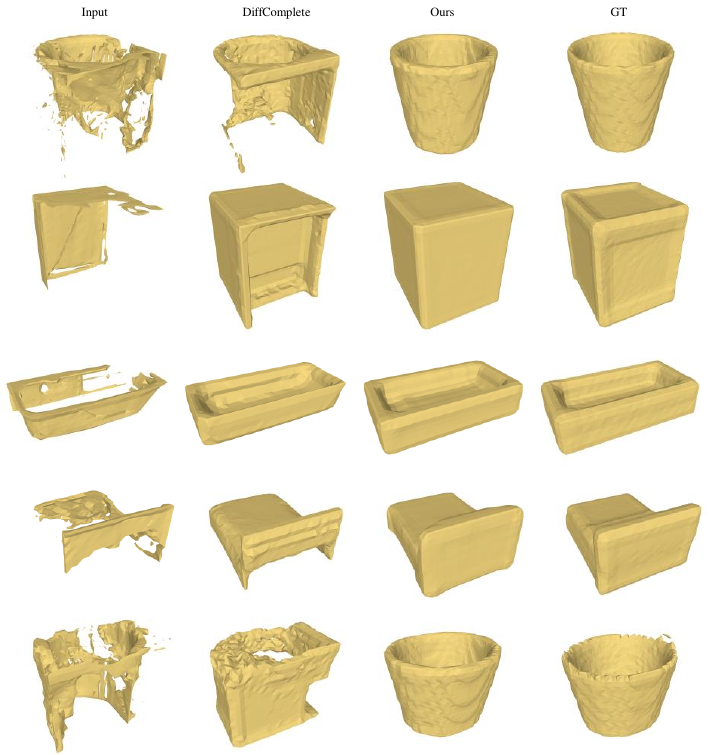}
\caption{Visual comparison with DiffComplete~\cite{chu2023diffcomplete} on Scannet~\cite{dai2017scannet}.}
\label{fig:scannet_more}
\end{figure*}

\subsection{Visualizations on Unseen Object Categories} As shown in Figure~\ref{fig:shapenet_more} and Figure~\ref{fig:scannet_more}, when applied to unseen object categories, our method outperforms DiffComplete~\cite{chu2023diffcomplete} in terms of both global consistency and local detail preservation. This demonstrates our method’s strong generalization capability, even when faced with real-world objects.

\section{Failure Cases and Limitations}
Although our method demonstrates promising results in 3D shape completion, several failure cases and inherent limitations indicate areas for further improvement.
Figure~\ref{fig:failure} presents several failure cases of our method. For known object categories (examples a and b), when the input is extremely sparse, the model sometimes fails to recover subtle geometric details—even though we introduce stochasticity to mitigate such cases—for instance, the intricate textures on cabinet surfaces. For unknown categories (examples c and d), the challenges are even more pronounced. In example (c), the model struggles to complete complex, cluttered lamp pendant structures, while in example (d), it erroneously predicts a sofa shape, despite the ground truth being a container. 
Overall, these failures highlight two key weaknesses: (1) the inability to complete fine details under extreme sparsity, and (2) limited generalization to highly diverse or irregular shapes. Beyond these challenges, our method also faces broader limitations related to latent space representation and computational efficiency.

Specifically, relying on a pre‑trained VQ‑VAE for latent encoding may limit our ability to generalize to highly complex or noisy shapes with intricate topology. Future work will investigate sparsity‑aware encoding schemes, end‑to‑end joint training of the VQ‑VAE and diffusion bridge, and more efficient latent representation learning techniques to improve robustness and reduce training overhead.

\end{document}